\documentclass[a4paper, 10pt, conference]{ieeeconf}      
\usepackage{FG2025}
\usepackage{amsmath}
\usepackage{comment}
\usepackage{graphicx}
 \usepackage{array}
 \usepackage{multirow}
 \usepackage{ragged2e}
 \usepackage{colortbl}
 \usepackage{xcolor}
 \usepackage{hyperref}
 \usepackage{amsfonts}

\IEEEoverridecommandlockouts                              
\def\FGPaperID{****} 

\title{\LARGE \bf
AffectSRNet : Facial Emotion-Aware Super-Resolution Network
}

\author{\parbox{16cm}{\centering
    {\large Syed Sameen Ahmad Rizvi$^*$, Soham Kumar$^*$, Aryan Seth$^*$, and Pratik Narang}\\
    {\normalsize Department of CSIS, Birla Institute of Technology \& Science, Pilani, RJ, India}\\
    {\normalsize Corresponding Author: Pratik Narang (e-mail: pratik.narang@pilani.bits-pilani.ac.in).}\\
    }
    \thanks{$^*$ These authors contributed equally.}}

\begin{document}

\ifFGfinal
\thispagestyle{empty}
\pagestyle{empty}
\else
\author{Anonymous FG2025 submission\\ Paper ID \FGPaperID \\}
\pagestyle{plain}
\fi
\maketitle

\begin{abstract}

Facial expression recognition (FER) systems in low-resolution settings face significant challenges in accurately identifying expressions due to the loss of fine-grained facial details. This limitation is especially problematic for applications like surveillance and mobile communications, where low image resolution is common and can compromise recognition accuracy. Traditional single-image face super-resolution (FSR) techniques, however, often fail to preserve the emotional intent of expressions, introducing distortions that obscure the original affective content. Given the inherently ill-posed nature of single-image super-resolution, a targeted approach is required to balance image quality enhancement with emotion retention. In this paper, we propose AffectSRNet, a novel emotion-aware super-resolution framework that reconstructs high-quality facial images from low-resolution inputs while maintaining the intensity and fidelity of facial expressions. Our method effectively bridges the gap between image resolution and expression accuracy by employing an expression-preserving loss function, specifically tailored for FER applications. Additionally, we introduce a new metric to assess emotion preservation in super-resolved images, providing a more nuanced evaluation of FER systems’ performance in low-resolution scenarios. Experimental results on standard datasets, including CelebA, FFHQ, and Helen, demonstrate that AffectSRNet outperforms existing FSR approaches in both visual quality and emotion fidelity, highlighting its potential for integration into practical FER applications. This work not only improves image clarity but also ensures that emotion-driven applications retain their core functionality in suboptimal resolution environments, paving the way for broader adoption in FER systems.
\end{abstract}

\section{INTRODUCTION}

Facial expressions are vital in indicating emotions, offering valuable clues about a person's emotional condition. The goal of automatic facial expression recognition is to create a dependable system capable of autonomously identifying and interpreting emotional signals based on facial characteristics. By accurately reading these emotional indicators, future human-computer interaction systems can become more user-friendly and attuned to human needs. Facial Expression Recognition (FER) lies at the intersection of two key fields: psychology and technology. In psychology, significant research has been devoted to comprehensively understanding and documenting how facial expressions align with emotional shifts. On the technological front, automation is enabled through the use of image processing techniques from computer vision combined with machine learning approaches.\\

As a result, super-resolution methods have achieved cross domain acceptability and enjoy a wide range of applications such as medical imaging \cite{sr3,sr4,sr1}, surveillance and security \cite{sr6,sr7}, aerial imaging \cite{sr8,sr9,sr10}, compressed image/video enhancement \cite{sr11}, action recognition \cite{sr12,sr13}, remote sensing \cite{sr14} , astronomical images \cite{sr15}, forensics \cite{sr16}, pose estimation \cite{sr17,sr18}, fingerprint and gait recognition \cite{sr19,sr20,sr21,sr22} and many more. Apart from improving the perceptual quality, it also helps in other deep learning-based computer vision tasks such as object detection and image segmentation.

Super-resolution (SR) is a classical problem in the field of image processing and computer vision as it is an ill-posed inverse problem, i.e., instead of a unique solution there can be multiple solutions for the same low-resolution image \cite{sr057}. Furthermore, the complexity of the problem is proportional to the upscaling factor, i.e. at higher upscaling rates, the retrieval of finer details is even more complex and results in the reproduction of wrong information.
\begin{center}
\begin{figure}
    \centering
    \includegraphics[scale=0.07]{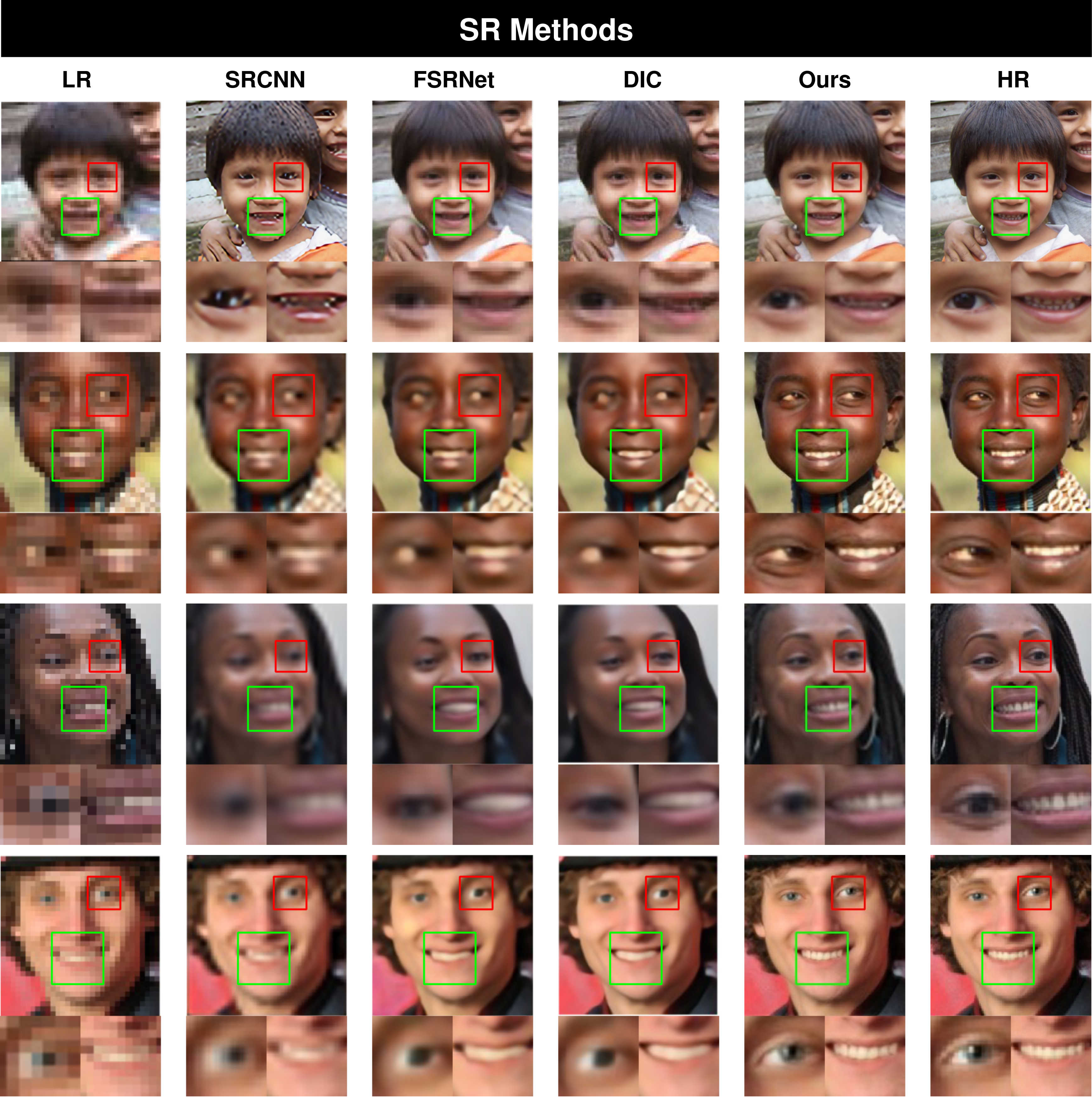}
    \caption{\label{top_image}Comparison of AffectSRNet with other super-resolution methods (SRCNN, FSRNet, DIC) on low-resolution facial images. AffectSRNet achieves superior image clarity and emotion fidelity, preserving fine facial details and expression accuracy.}
\end{figure}
\end{center}

Real-world in-the-wild applications of FER involve education, security, marketing and software development which often requires the deep FER application to process low-resolution video footage. A lot of research has explored face super-resolution (FSR), which includes the likes of identity-aware FSR and attribute constraint FSR. However, facial expressions get distorted while upsampling an image from low to high resolutions. This subdomain of expression-aware FSR, is under-explored and this research attempts to tackle this problem.
\subsection{Our Contributions}
In this paper, we introduce a novel emotion-aware face super-resolution pipeline designed to preserve facial expressions while upscaling low-resolution facial images. This approach enables the practical deployment of facial expression recognition (FER) applications on low-resolution image sequences. The key contributions of this work include:
\begin{itemize}
    \item We propose a novel technique based on graph embeddings for emotion-aware single-image face super-resolution, which upscales facial images while maintaining the intensity of facial expressions.
    \item We propose a novel metric to assess the accuracy of emotion-aware face super-resolution.
    \item We provide comprehensive quantitative as well as quantitative comparisons on state-of-the-art FSR methods on CelebA \cite{celebA}, FFHQ \cite{FFHQ}, and Helen \cite{helen} datasets. 
    \item We formulate an extensible loss function, which can be integrated into existing FSR networks through fine-tuning.
\end{itemize}
\section{Related Works}
With the evolution of deep learning paradigm, notable advances have been done in image super-resolution. Exhaustive surveys \cite{sr056,sr057,sr058} have been done to summarize the prominent works done in deep learning based image super-resolution. In this section, we mention few notable works discussed in the literature.

\subsection{Supervised Super-resolution}
Due to the ill-posed nature of image super-resolution, the primary challenge lies in determining how to effectively perform upsampling—converting low-resolution (LR) input into high-resolution (HR) output. Despite the wide range of model architectures, they can generally be grouped into four primary frameworks.

\subsubsection{Pre-Upsampling Super-resolution}
Given the difficulty of directly learning the mapping from low-dimensional to high-dimensional space, a practical solution involves utilizing traditional upsampling techniques to generate higher-resolution images, followed by refinement using deep neural networks (DNNs). Dong et al. \cite{s1,s2} were the first to adopt this pre-upsampling framework, proposing SRCNN. In this approach, LR images are initially upsampled to coarse HR images of the desired size using standard methods such as bicubic interpolation. Afterward, deep CNNs are employed to enhance the fine details. Since the upsampling step, which is computationally demanding, has already been carried out, the CNNs only need to refine the upsampled images, thus lowering the complexity of the learning task. Additionally, these models can handle interpolated images of arbitrary sizes and scaling factors, offering performance comparable to single-scale SR models \cite{s3}. This framework has gained widespread popularity \cite{s4,s5,s6,s7}, with variations primarily in the design of the posterior model and learning strategies. However, pre-defined upsampling can lead to side effects, such as noise amplification and blurring. Furthermore, performing operations in high-dimensional space incurs higher computational and memory costs compared to other frameworks \cite{s8,s9}.

\subsubsection{Post-Upsampling Super-resolution}
To enhance computational efficiency and maximize the benefits of deep learning for resolution enhancement, researchers have proposed conducting most computations in low-dimensional space, with end-to-end learnable upsampling layers applied at the network’s output. This framework, called post-upsampling SR, avoids the pre-defined upsampling step by feeding the LR images directly into deep CNNs, leaving the resolution increase for the final stage. This design significantly reduces both computational and spatial complexity since the feature extraction—which is computationally intensive—occurs in low-dimensional space, and resolution is only increased during the final stage. As a result, this approach has become one of the most widely adopted frameworks \cite{s10,s11,s12,s13}. The differences across models within this framework largely lie in the design of learnable upsampling layers, the preceding CNN structures, and the learning strategies used.
\subsubsection{Progressive Upsampling Super-resolution}
While the post-upsampling framework offers a significant reduction in computational cost, it still faces challenges. Upsampling is performed in a single step, which increases the difficulty of learning when handling large scaling factors (e.g., 4x, 8x). Moreover, separate models must be trained for different scaling factors, which limits flexibility for multi-scale SR tasks. To overcome these limitations, the progressive upsampling framework was introduced by Laplacian pyramid SR network (LapSRN) \cite{s14}. This framework incrementally reconstructs HR images through a cascade of CNNs, with each stage progressively upsampling and refining the images. Other models, such as MS-LapSRN \cite{s15} and progressive SR (ProSR) \cite{s16}, have adopted this framework and achieved high performance. In contrast to LapSRN and MS-LapSRN, which treat intermediate reconstructed images as ``base images" for subsequent modules, ProSR maintains a central information stream, reconstructing intermediate-resolution images using separate heads.

\subsubsection{Iterative Up-and-Down Sampling Super-resolution}
To better model the interdependence between LR and HR image pairs, an efficient iterative refinement process known as back-projection \cite{s17} has been incorporated into super-resolution models \cite{s18}. This framework, referred to as iterative up-and-down sampling SR, iteratively applies back-projection refinement, computing reconstruction errors and using them to adjust HR image intensities. Haris et al. \cite{s19} introduced DBPN, which alternates between upsampling and downsampling layers, ultimately reconstructing the HR output from a series of intermediate reconstructions. Likewise, SRFBN \cite{s20} employs an iterative feedback mechanism with dense skip connections to enhance representational learning, while RBPN \cite{s21} applies this framework to video super-resolution, combining context from multiple video frames to produce recurrent HR outputs through a back-projection module.

\subsection{Unsupervised Super-resolution}
Most super-resolution methods to date have focused on supervised learning, relying on matched LR-HR image pairs for training. However, in practice, it is often difficult to obtain images of the same scene at different resolutions, leading to datasets where LR images are generated through predefined degradation processes applied to HR images. As a result, the models trained using these datasets may not generalize well to real-world scenarios.

\subsubsection{Zero-Shot Super-resolution}
Acknowledging that the internal image statistics within a single image provide enough information for super-resolution, Shocher et al. \cite{s22} proposed zero-shot super-resolution (ZSSR). ZSSR addresses unsupervised SR by training image-specific SR networks during the testing phase, rather than relying on a generic model trained on large external datasets. Specifically, a degradation kernel is estimated from a single image \cite{s23}, and a small dataset is created by applying degradations and augmentations to the image using various scaling factors. A CNN is then trained on this small dataset for SR, leveraging the internal cross-scale recurrence found within the image. This approach significantly outperforms prior methods, especially in non-ideal conditions (e.g., noisy, blurry, or compressed images), with improvements of up to 1 dB for estimated kernels and 2 dB for known kernels. However, the method requires training a different network for each test image, resulting in longer inference times.

\subsubsection{Weakly-Supervised Super-resolution}
Because predefined degradation is suboptimal, learning degradation from unpaired LR-HR datasets presents a promising alternative. Bulat et al. \cite{s24} proposed a two-stage process where an HR-to-LR GAN is trained using unpaired LR-HR images to learn degradation, followed by training an LR-to-HR GAN for SR using the generated LR-HR image pairs. Specifically, the HR-to-LR GAN generates LR images from HR inputs that are required to match not only the LR images produced through downscaling (e.g., via average pooling) but also the distribution of real LR images. Once trained, this generator is used to create LR-HR pairs. The LR-to-HR GAN (which serves as the SR model) then uses these generated LR images to predict HR outputs. Inspired by CycleGAN \cite{s25}, Yuan et al. \cite{s26} proposed the cycle-in-cycle SR network (CinCGAN), which consists of four generators and two discriminators forming two CycleGANs that model noisy LR to clean LR and clean LR to clean HR mappings. In the first CycleGAN, the noisy LR image is passed through a generator that outputs an image matching the distribution of real clean LR images, which is then used as input for the second generator to recover the original HR image.

\subsubsection{Deep Image Prior Super-resolution}
Prior-guided FSR methods consistently harness the potential of facial priors to enhance face super-resolution \cite{fsr21,fsr22,fsr23}. Chen et al. \cite{fsr24} carried out face reconstruction from a coarse to fine level, utilizing prior information specific to faces. Kim et al. \cite{fsr25} created multiple multi-resolution facial images through a progressive training approach. Ma et al. \cite{fsr26} integrated facial super-resolution (FSR) with landmark estimation, using an iterative and recursive method for face reconstruction. Yin et al. \cite{fsr27} introduced a multi-task framework that simultaneously learns face landmarks and super-resolution, where the tasks mutually support each other. Wang et al. \cite{fsr28} introduced an innovative dual closed-loop network (DCLNet) based on CNNs to reduce the potential mapping space. Li et al. \cite{fsr29} developed a five-branch network focusing on five key regions of the human face for face hallucination. Kalarot et al. \cite{fsr30} improved FSR outcomes using facial component attention maps. Similarly, methods leveraging face semantic information and heat maps are increasingly being explored. For addressing higher magnification factors, Liu et al. \cite{fsr31} proposed an FSR method that incorporates face parsing maps. Zhao and Zhang \cite{fsr32} introduced adaptive FSR using face semantic attention. Zhang et al. \cite{fsr33} combined a deep neural network (DNN) with both a face image super-resolution branch and a semantic face parsing branch. Wang et al. \cite{fsr34} designed a novel heat map-aware convolution utilizing spatially variant kernels, rather than the conventional spatially shared kernel, to restore different facial regions. Additionally, Wang et al. \cite{fsr35} created a new FSR network that employs resolved maps to directly extract face prior information from low-resolution images for subsequent use.
\subsection{Face Super-Resolution -- A Domain Specific SR }
Face super-resolution is a domain-specific problem within the realm of super-resolution. FSR has attracted increased attention in research communities and achieved significant advancements. In literature, FSR has broadly been categorized into two types on the basis of their approach to upsample a given image. These include Network Architecture Design-based FSR and Facial Prior-guided FSR. We shall discuss both of these briefly.\\
\subsubsection{Network Architecture Design-based FSR}
Due to the rapid advancements in deep convolutional neural networks (CNNs), deep learning techniques have been extensively applied to computer vision tasks \cite{fsr1,fsr2,fsr3,fsr4}. Over the past few years, there has been a notable increase in leveraging deep CNN models for face super-resolution (FSR) \cite{fsr5}, resulting in continuous improvements in performance. Early deep learning approaches to FSR primarily focused on designing efficient network architectures. For example, Zhou et al. \cite{fsr6} introduced a bi-channel network to extract informative features, which were then fused for FSR. Huang et al. \cite{fsr8} proposed a wavelet-based CNN method designed to ultra-resolve extremely low-resolution facial images. Drawing inspiration from attention mechanisms \cite{fsr9,fsr10}, Chen et al. [24] created face attention units specifically tailored for FSR. To improve feature representation, Lu et al. \cite{fsr11} introduced the global-local split-attention network, which applies local attention to groups of feature maps while achieving global attention. Jiang et al. \cite{fsr12}, without relying on extra prior knowledge, developed a dual-path deep fusion network consisting of two distinct branches to accomplish face image super-resolution. Chen et al. \cite{fsr13} proposed a lightweight single-image super-resolution network that integrates multi-level features to tackle the typical issues of blurred image edges, the rigid selection of convolution kernel sizes, and slow convergence during training caused by redundant network structures in existing image super-resolution algorithms. In \cite{fsr14}, authors propose an innovative deep hybrid feature-based attention model specifically designed for FSR.

Drawing inspiration from generative adversarial networks (GAN), Yu et al. \cite{fsr7} developed the first GAN-based FSR method, URDGN, to reconstruct realistic face images. ATFaceGAN \cite{fsr15} adopts a dual-path training approach to improve facial images, while HiFaceGAN \cite{fsr16} presents a suppression module specifically designed to enhance high-frequency details effectively. To address the shortcomings of traditional GANs in FSR, a supervised pixel-wise GAN is specifically designed to enhance low-resolution face images \cite{fsr17}. However, training the discriminator to recognize the entire face image poses challenges. To overcome this, Dou et al. \cite{fsr18} break down the face images into different components, allowing the discriminator to gradually learn these parts. The approach in \cite{fsr19} introduces a generative and controllable FSR framework, while ECSRNet \cite{fsr20} directly captures both low- and high-frequency details using a progressive asymmetric architecture to perform face hallucination. 
\subsubsection{Facial Prior Guided FSR}
Prior-guided FSR methods consistently harness the potential of facial priors to enhance face super-resolution \cite{fsr21,fsr22,fsr23}. Chen et al. \cite{fsr24} carried out face reconstruction from a coarse to fine level, utilizing prior information specific to faces. Kim et al. \cite{fsr25} created multiple multi-resolution facial images through a progressive training approach. Ma et al. \cite{fsr26} integrated facial super-resolution (FSR) with landmark estimation, using an iterative and recursive method for face reconstruction. Yin et al. \cite{fsr27} introduced a multi-task framework that simultaneously learns face landmarks and super-resolution, where the tasks mutually support each other. Wang et al. \cite{fsr28} introduced an innovative dual closed-loop network (DCLNet) based on CNNs to reduce the potential mapping space. Li et al. \cite{fsr29} developed a five-branch network focusing on five key regions of the human face for face hallucination. Kalarot et al. \cite{fsr30} improved FSR outcomes using facial component attention maps. Similarly, methods leveraging face semantic information and heat maps are increasingly being explored. For addressing higher magnification factors, Liu et al. \cite{fsr31} proposed an FSR method that incorporates face parsing maps. Zhao and Zhang \cite{fsr32} introduced adaptive FSR using face semantic attention. Zhang et al. \cite{fsr33} combined a deep neural network (DNN) with both a face image super-resolution branch and a semantic face parsing branch. Wang et al. \cite{fsr34} designed a novel heat map-aware convolution utilizing spatially variant kernels, rather than the conventional spatially shared kernel, to restore different facial regions. Additionally, Wang et al. \cite{fsr35} created a new FSR network that employs resolved maps to directly extract face prior information from low-resolution images for subsequent use.

\begin{figure*}
    \centering
    \includegraphics[scale=0.105]{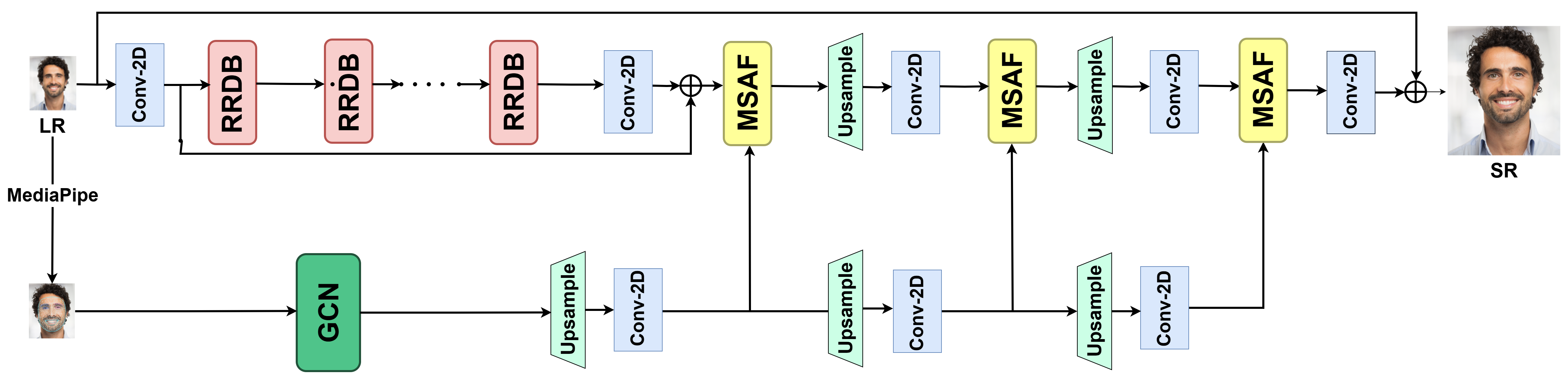}
    \caption{\label{arc} The network architecture of AffectSRNet. The super-resolution backbone consists of the RRDB and upsampling blocks from ESRGAN\cite{fsr36_esrgan}. Facial landmarks extracted with Mediapipe\cite{Mediapipe} are passed through GCN block to get graph embeddings. These are integrated into the super-resolution backbone, with MSAF block performing cross-modal fusion.}
   
\end{figure*}

\section{Methodology}
This section outlines the architectural design and operational mechanism of the proposed network. Figure \ref{arc} illustrates the overall architecture of AffectSRNet, highlighting how facial landmark priors are integrated into the super-resolution backbone to preserve facial structural details. Facial landmarks are first extracted from the low-resolution image using a pretrained network. These landmarks are then processed through a Graph Convolutional Network (GCN) block to generate graph embeddings. The graph embeddings are integrated into the super-resolution backbone through an attention-based multimodal fusion block, infusing structural information of face into the process, thereby preserving facial expressions.\\
Our pipeline for training is to compare original images of full size and their super-resolved counterparts, produced by images which have been downsampled.

\subsection{Evaluating Emotion Consistency}
We formulate an Emotion Consistency Loss, which is used to measure how strongly an emotion is retained across the super-resolution.\\
It is important to note that these are used as metrics and not used during training. This is due to this loss not being convex, making it difficult to optimize over effectively.\\

We start by defining notation: Input image (lower dimensional) is denoted by $X$, Target Image (higher dimensional, which is downsampled to form $X$) is denoted by $Y$, the super resolution model architecture is denoted by $M$, and the super-resolved image is denoted by $I_M$
The notion of consistency in our problem setting arises from our ability to accurately predict an emotion, therefore it is important that our metric reflects this change. There are two metric formulations that we use to describe this, first we focus on the change in confidence between two classification models in the original and super-resolved images. To do this, we use the Histogram Loss between two output distributions for histograms created by the confidence scores of a Facial Expression Recognition model for all samples in the test set. This is done to examine how different the predictive estimates for the original and super-resolved images are.

The Histogram Loss \( L_H \) between the confidence scores of the original image \( X \) and the super-resolved image \( I_M \) is given by:

\[
L_H = \sum_{c=1}^{C} \left( H(p_X^c) - H(p_{I_M}^c) \right)^2
\]

where \( H(p_X^c) \) and \( H(p_{I_M}^c) \) are the histograms of confidence scores for class \( c \) in the original and super-resolved images, respectively.

The second metric we use for evaluating consistency is the average difference in predictive confidence for all samples by an auxiliary Facial Expression Recognition model. We model predictive confidence as the entropy of the logits of the outputs.

The entropy for a given image \( X \) is defined as:

\[
H(p_X) = -\sum_{c=1}^{C} p_X^c \log p_X^c
\]

where \( p_X^c \) is the probability for class \( c \) in the output distribution for image \( X \), and \( C \) is the total number of classes.

The average difference in predictive confidence between the original image \( X \) and the super-resolved image \( I_M \) across all samples is given by:

\[
L_{\text{conf}} = \frac{1}{N} \sum_{i=1}^{N} \left| H(p_{X_i}) - H(p_{I_{M_i}}) \right|
\]

where \( N \) is the number of samples in the test set, and \( H(p_{X_i}) \) and \( H(p_{I_{M_i}}) \) are the entropies of the logits for the \( i \)-th original and super-resolved images, respectively.

The final Emotion Consistency Metric (ECM) is defined as a log-weighted combination of the Histogram Loss $L_H$ and the average difference in predictive confidence between the original image $X$ and the super-resolved image $I_M$ ($L_\text{conf}$):

\[
\text{ECM} = \alpha L_H + \log( L_\text{conf})
\]
where $\alpha = 0.5$.

In practice, this auxiliary model is designed to be a plugin, ideally trained on a variety of datasets. In our experiments, we use DDAMFN++ \cite{ddamfn} for this task.

\subsection{Super-resolution Backbone}
The super-resolution backbone network utilizes the Residual-in-Residual Dense Block (RRDB) architecture, originally introduced in ESRGAN \cite{fsr36_esrgan}. RRDB combines multi-level residual networks with dense connections, implementing residual learning across multiple layers to create a nested residual structure.

Each Residual Dense Block (RDB) within RRDB consists of five stacked convolutional layers, where each layer has a residual connection to every other layer in the block. An RRDB is composed of three sequential RDBs, with an overarching residual connection across the entire block.

The backbone network is constructed by stacking multiple RRDB blocks. The low-resolution input image is first passed through an initial convolutional layer to extract features. This is followed by a trunk formed by several RRDB blocks and another convolutional layer. The network then includes upsampling layers, and concludes with the final convolutional layers that generate the super-resolution output. A residual connection is added from the input the final output.

\subsection{Graph Convolutional Network block}
\begin{center}
\begin{figure}
    \centering
    \includegraphics[scale=0.20]{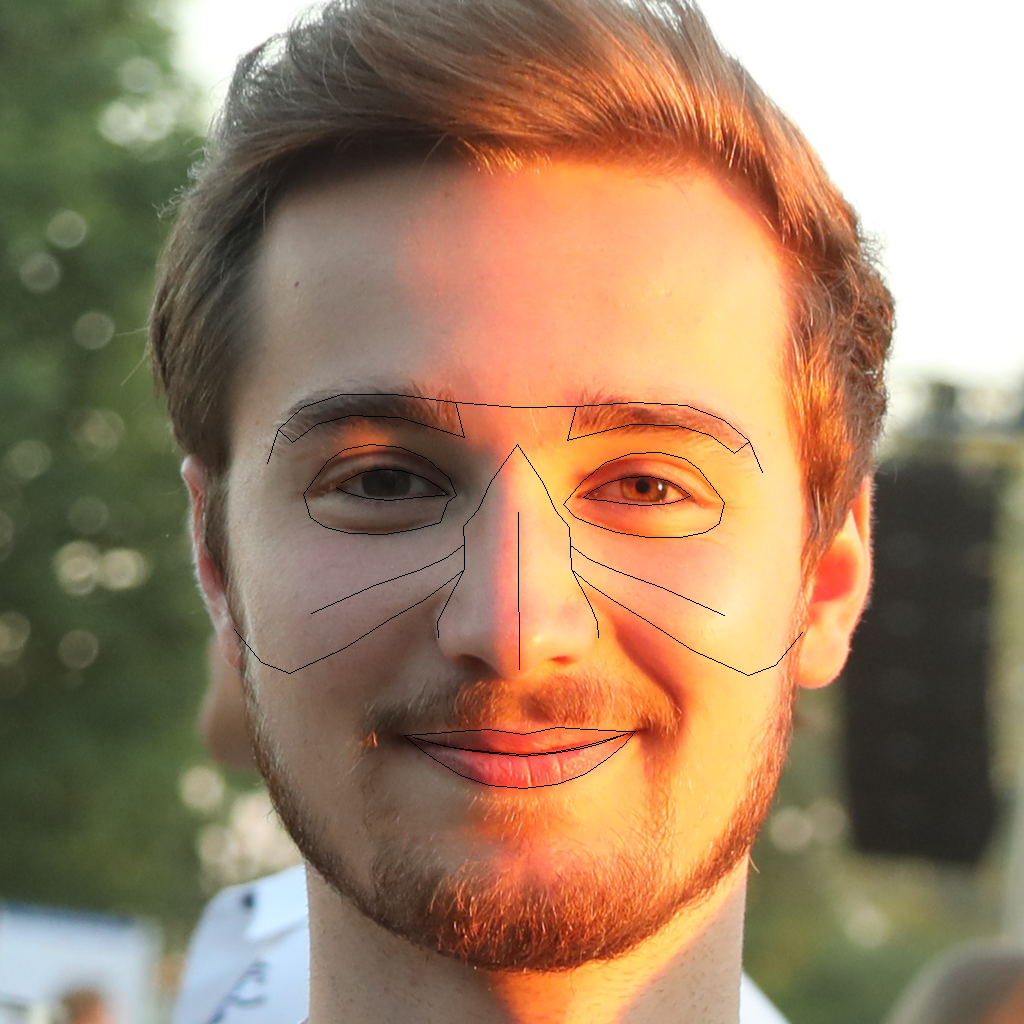}
    \caption{\label{edges}The edges are defined as illustrated to preserve the spatial dependence of the features important for facial expression.}
   
\end{figure}
\end{center}

The proposed architecture uses facial landmark priors to preserve facial expressions of both low-resolution in the super-resolved images. A total of 478 3D facial landmarks, extracted using Mediapipe \cite{Mediapipe}, are processed through a Graph Convolutional Network (GCN) \cite{gcn} block to generate graph embeddings. These embeddings encode the relative orientation of key facial landmarks in 3D space, preserving the facial expressions. It has been shown that Facial Prior Knowledge is \textit{significant} for facial super-resolution \cite{fsr24}. Edges between the 3D facial landmarks are manually defined to connect key regions such as the eyes, lips, and cheeks, preserving their relative positions and structural relationships, as illustrated in Fig.\ref{edges}. The extracted graph embeddings are upscaled and merged in the super resolution backbone thrice Fig.\ref{arc}. The GCN block contains stacked GCN layers.
\subsection{Multimodal Split Attention Fusion Block}

The Multimodal Split Attention Fusion (MSAF) \cite{msaf} block fuses the graph embeddings from the GCN block and the intermediate embeddings in the Super-Resolution backbone. The MSAF module operates on feature maps from different modalities, generating optimized feature maps through a process of channel-wise splitting, joining, and highlighting. Initially, each feature map is split into equal-channel feature blocks. The collection of feature blocks corresponding to modality \( m \) is denoted as \( B_m \), where \( |B_m| = \lceil C_m / C \rceil \), and \( m \in \{ 1, \dots, M \} \). The \( i \)-th feature block within \( B_m \) is denoted by \( B^i_m \), where \( i \in \{ 1, \dots, |B_m| \} \). 

The feature blocks of modality \( m \) are combined into a shared representation \( D_m \) by calculating the element-wise sum \( S_m \) over \( B_m \), followed by global average pooling across the spatial dimensions:

\[
D_m(c) = \frac{1}{\prod_{i=1}^{K} N_i} \sum_{(n_1,\dots,n_K)} S_m(n_1, n_2, \dots, n_K, c)
\]

Each channel descriptor is summarized into a feature vector of length \( C \). To derive a multimodal representation, the element-wise sum of all modality descriptors \( \{D_1, \dots, D_M\} \) is computed, forming the multimodal channel descriptor \( G \).

The channel-wise dependencies are modelled through a fully connected layer with a reduction factor \( r \), followed by batch normalization and ReLU. This maps \( G \) to the joint representation \( Z \in \mathbb{R}^{C'} \), where \( C' = \left\lfloor \frac{C}{r} \right\rfloor \):

\[
Z = W_Z G + b_Z
\]

where \( W_Z \in \mathbb{R}^{C' \times C} \) and \( b_Z \in \mathbb{R}^{C'} \).
This forms a shared representation that captures the multimodal global context. This context is used to generate per-channel block-wise attention, highlighting relevant features across modalities.

\subsection{Loss Functions}
\subsubsection{L1 Loss}
We use L1-norm as the pixel level loss. It mainly helps to constrain the low level information in the outputs especially color.
$\mathcal{L}_{pix}^h=\frac1N\sum_{i=1}^N\|I_{SR}^i-I_{HR}^i\|_1$

\subsubsection{Perceptual Loss}
Style Reconstruction Loss. The feature reconstruction loss penalizes the output image $\hat{y}$ when it deviates in content from the target $y.$ We also wish to penalize differences in style: colors, textures, common patterns, etc. To achieve this effect, Gatys $et$ $al$ [9,10] propose the following $style\textit{ reconstruction loss. }$
As above, let $\phi_j(x)$ be the activations at the $j$th layer of the network $\phi$ fon the input $x$, which is a feature map of shape $C_j\times H_j\times W_j.$ Define the Gram $matrix~G_{j}^{\phi}(x)$ to be the $C_j\times C_j$ matrix whose elements are given by

$$G_{j}^{\phi}(x)_{c,c'}=\frac{1}{C_{j}H_{j}W_{j}}\sum_{h=1}^{H_{j}}\sum_{w=1}^{W_{j}}\phi_{j}(x)_{h,w,c}\phi_{j}(x)_{h,w,c'}.$$

\subsubsection{L2 Loss for Node Embeddings}
Let $G_1$ and $G_2$ be two graphs, and let $\mathbf{z}_v^{G_1}$ and $\mathbf{z}_v^{G_2}$ be the embeddings of node $v$ in graphs $G_1$ and $G_2$, respectively. The L2 loss between the node embeddings is:

\begin{equation}
L_{\text{L2}} = \sum_{v \in V} \|\mathbf{z}_v^{G_1} - \mathbf{z}_v^{G_2}\|_2^2
\end{equation}

The final loss is calculated as:
\begin{equation}
L_{\text{total}} = k_1 \cdot L_{\text{pix}}^h + k_2\cdot L_{\text{hist}} + k_3 \cdot G_{j}^{\phi}(x)_{c,c'} + k_4 \cdot L_{\text{L2}}
\end{equation}
Where $k_1$, $k_2$, $k_3$ and $k_4$ are hyperparameters with values $k_1$ = 1, $k_2$ = 20, $k_3$ = 50 and $k_4$ = 0.1.

\begin{table*}
\centering
\caption{\label{4x}Quantitative assessment of different face super-resolution (FSR) techniques on the CelebA\cite{celebA}.Helen \cite{helen} and FFHQ\cite{FFHQ}, for an upsampling factor of 4.}
\begin{tabular}{|c|c|ccccccc|}
\hline
\multirow{2}{*}{\textbf{Datasets}} & \multirow{2}{*}{\textbf{Metric}} & \multicolumn{7}{c|}{\textbf{Methods}} \\ 
\cline{3-9}
 &  & \textbf{Bicubic} & \textbf{SRCNN\cite{SRCNN}} & \textbf{EDSR\cite{EDSR}} & \textbf{FSRNET\cite{FSRNet}} & \textbf{DIC\cite{DIC}} & \textbf{SPARNET\cite{SPARNet}} & \textbf{Ours} \\ 
\hline
\multirow{4}{*}{\textbf{CelebA\cite{celebA}}} & PSNR~↑ & 27.48 & 28.04 & 31.45 & 31.46 & 31.53 & \textbf{31.71} & 31.68 \\
 & SSIM~↑ & 0.8166 & 0.8369 & 0.9095 & 0.9084 & 0.9107 & 0.9129 & \textbf{0.9141}\\
 & LPIPS ↓ & 0.3589 & 0.1599 & 0.0518 & 0.0519 & 0.0532 & 0.0476 &  \textbf{0.0394}\\
 & ECM~↓ & 16.88 & 15.78 & 13.26 & 13.05 & 12.67 & 12.12 &  \textbf{10.11}\\ 
\hline
\multirow{4}{*}{\textbf{Helen\cite{helen}}} & PSNR~↑ & 28.22 & 28.77 & 31.87 & 31.93 & \textbf{31.98} & \textbf{31.98} & 31.94 \\
 & SSIM ↑ & 0.6628 & 0.8730 & 0.9286 & 0.9283 & 0.9303 & 0.9300 & \textbf{0.9320}\\
 & LPIPS ↓ & 0.1771 & 0.556 & 0.0574 & 0.0543 & 0.0576 & 0.0592 &  \textbf{0.0522}\\
 & ECM~↓ & 15.97 & 15.24 & 13.22 & 13.15 & 12.06 & 11.23 &  \textbf{9.89}\\ 
\hline
\multirow{4}{*}{\textbf{FFHQ\cite{FFHQ}}} & PSNR~↑ & 29.82 & 32.65 & 31.9 & 31.94 & - & 32.36 & \textbf{32.42} \\
 & SSIM ↑ & 0.8459 & 0.8980 & 0.9161 & 0.9155 & - & 0.8933 & \textbf{0.928 }\\
 & LPIPS ↓ & 0.3361 & 0.1720 & 0.0502 & 0.0498 & - & 0.1878 &  \textbf{0.1260}\\
 & ECM~↓ & 16.02 & 15.03 & 13.44 & 12.34 & - & 10.87 &  \textbf{9.64}\\
\hline
\end{tabular}
\end{table*}

\begin{table*}
\centering
\caption{\label{8x}Quantitative assessment of different face super-resolution (FSR) techniques on the CelebA\cite{celebA}.Helen \cite{helen} and FFHQ\cite{FFHQ}, for an upsampling factor of 8.}
\begin{tabular}{|c|c|ccccccc|}
\hline
\multirow{2}{*}{\textbf{Datasets}} & \multirow{2}{*}{\textbf{Metric}} & \multicolumn{7}{c|}{\textbf{Methods}} \\ 
\cline{3-9}
 &  & \textbf{Bicubic} & \textbf{SRCNN\cite{SRCNN}} & \textbf{EDSR\cite{EDSR}} & \textbf{FSRNET\cite{FSRNet}} & \textbf{DIC\cite{DIC}} & \textbf{SPARNET\cite{SPARNet}} & \textbf{Ours} \\ 
\hline
\multirow{4}{*}{\textbf{CelebA\cite{celebA}}} & PSNR~↑ &  23.58 & 23.93 & 26.84 & 26.66 & 27.37 & \textbf{27.42} & 27.39 \\
 & SSIM~↑ & 0.6258 & 0.6348 & 0.7787 & 0.7714 & 0.8022 & 0.8036 &  \textbf{0.8124}\\
 & LPIPS ↓ & 0.6290 & 0.2559 & 0.1159 & 0.1098 & 0.0920 & 0.0891 &  \textbf{0.0768}\\
 & ECM~↓ &  17.23&  15.83&  13.85&  13.66&  12.94&  12.48&  \textbf{11.26}\\ 
\hline
\multirow{4}{*}{\textbf{Helen\cite{helen}}} &  PSNR~↑ & 23.88 & 24.27 & 26.60 & 26.43 & 26.94 & 2\textbf{6.95} &  26.68\\
 & SSIM~↑ & 0.6628 &  0.6770 & 0.7851 & 0.7799 & 0.8026 & 0.8029 &  \textbf{0.8086}\\
 & LPIPS~↓ & 0.2560 & 0.2430 & 0.1400 & 0.1356 & 0.1144 & 0.1169 &  \textbf{0.1023}\\
 & ECM~↓ &  16.34&  16.02&  13.87&  13.65&  12.97&  11.87&  \textbf{10.97}\\ 
\hline
\multirow{4}{*}{\textbf{FFHQ\cite{FFHQ}}} &PSNR~↑ & 25.99 & 28.17 & 27.72 & 27.78 & -& 28.20 &  \textbf{28.24}\\
 & SSIM~↑ & 0.7313 & 0.7932 & 0.7841 & 0.7839 & -& 0.7965 &  \textbf{0.7986}\\
 & LPIPS~↓ & 0.5594 & 0.3329 & 0.3554 & 0.3552 & -& 0.3355 &  \textbf{0.3234}\\
 & ECM~↓ &  16.55&  15.49&  14.13&  12.82&  -&  11.77&  \textbf{10.54}\\
\hline
\end{tabular}
\end{table*}

\begin{table*}
\centering
\caption{\label{ablnsr} Component wise ablation study of  AffectSRNet}
\begin{tabular}{|c|ccc|ccc|ccc|}
\hline
\multirow{2}{*}{Component} & \multicolumn{3}{c|}{CelebA} & \multicolumn{3}{c|}{Helen} & \multicolumn{3}{c|}{FFHQ} \\ 
\cline{2-10}
 & PSNR↑ & SSIM↑ & ECM↓ & PSNR↑ & SSIM↑ & ECM↓& PSNR↑ & SSIM↑ & ECM↓\\ 
\hline
RRDB &  25.93&  0.7811&  14.12&  24.36&  0.7232&  13.89&  26.25&  0.7112&  14.54\\ 
\hline
RRDB + Instance Norm &  26.54&  0.7924&  13.89&  26.22&  0.7826&  13.11&  27.13&  0.7621&  13.76\\ 
\hline
\textbf{RRDB + Instance Norm + MSAF + GCN (Ours)} &  27.46&  0.8124&  11.26&  26.68&  0.8086&  10.97&  28.24&  0.7986&  10.54\\
\hline
\end{tabular}
\end{table*}

\section{Experimentation, Results and Analysis}
\subsection{Dataset and Metrics}
This study utilizes three well-known facial datasets: FFHQ \cite{FFHQ}, CelebA \cite{celebA}  and Helen \cite{helen}. We use Mediapipe\cite{Mediapipe} to extract 478 facial landmarks. For generating low-resolution (LR) face images, the ground truth is downsampled to 32×32 and 16×16 pixels using bicubic interpolation, representing 4× and 8× FSR tasks, respectively. During training, the CelebA, FFHQ, and Helen datasets are employed. For testing, we evaluate model performance using 1000 images from FFHQ and CelebA, and 50 images from Helen. The evaluation metrics include PSNR, SSIM \cite{SSIM}, LPIPS \cite{LPIPS} and the Emotion Consistency Metric (ECM).

\subsection{Implementation Details}
In the super-resolution backbone we use 8 blocks, and the GCN block comprises 4 consecutive GCN layers. Skip connections are done through bicubic interpolation of the input image. Bicubic interpolation is similarly used in all the upsample + convolution modules. The training process uses pre-trained ESRGAN weights for the super-resolution backbone. We utilize the Adam optimizer with a learning rate ranging from 1e-3 to 1e-5. All experiments are conducted using PyTorch \cite{PyTorch} on an Nvidia RTX A6000 GPU.

\subsection{Comparitive Study}
\subsubsection{Quantitative Results}
To evaluate the effectiveness of our approach in facial super-resolution, especially in preserving emotions, we compare it with several leading methods. These include two CNN-based general image super-resolution techniques, SRCNN \cite{SRCNN} and EDSR \cite{EDSR}, along with four facial super-resolution methods: FSRNet \cite{FSRNet}, DIC \cite{DIC}, and SPARNet \cite{SPARNet}. Bicubic interpolation is also used as a baseline. Table \ref{4x} and \ref{8x} presents the quantitative results. For fairness, all models are trained and tested on the same dataset. Across all three datasets, our method delivers results comparable to the state-of-the-art in PSNR, SSIM, and LPIPS while outperforming others in ECM. Notably, SRCNN and EDSR, being non-face-specific, struggle to accurately reconstruct facial images.
\begin{figure*}
    \centering
    \includegraphics [width=16cm,height=23cm]{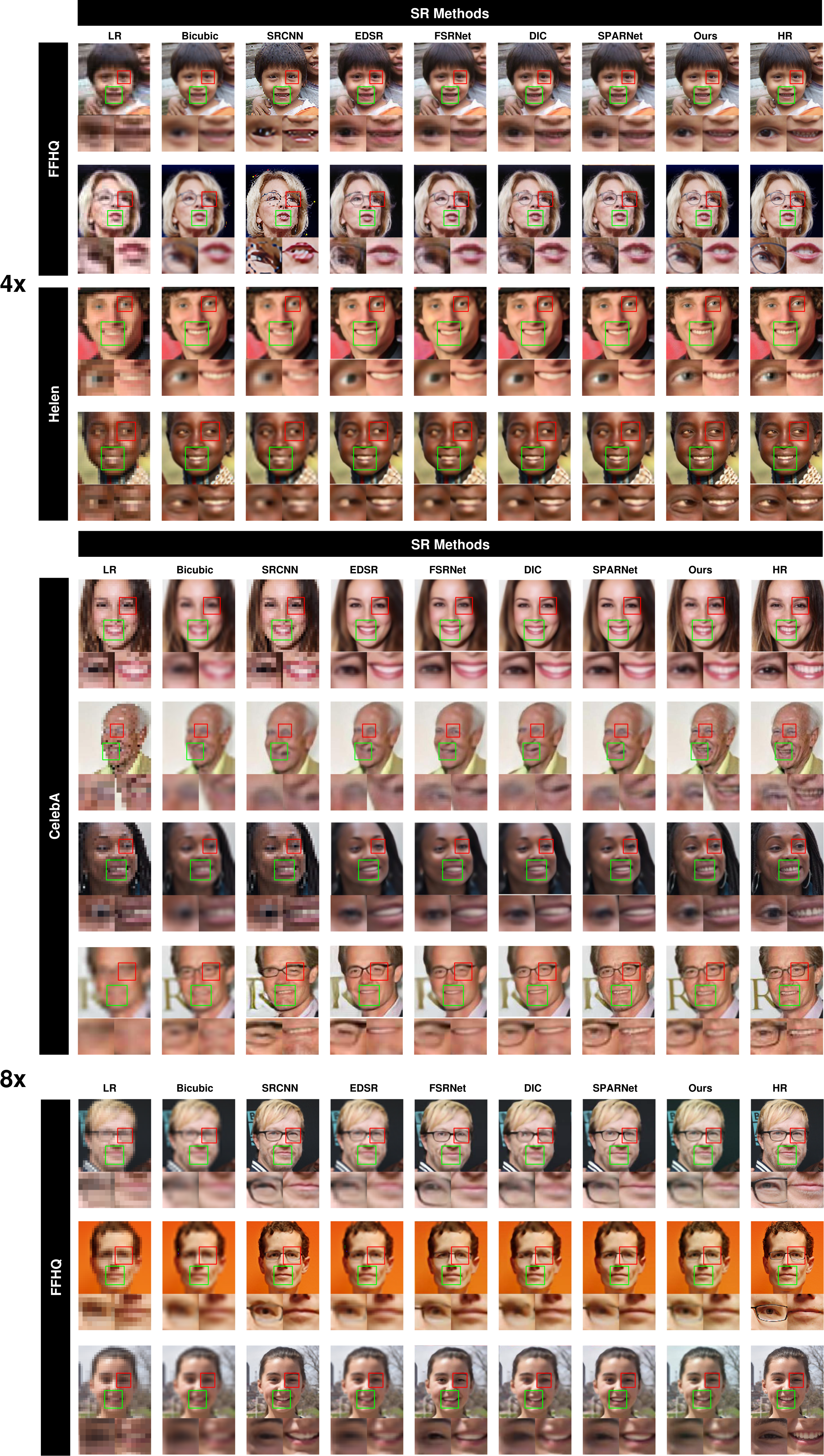}
    \caption{\label{visual}The figure shows a visual comparison of leading methods applied to the Helen\cite{helen}, FFHQ \cite{FFHQ} and CelebA\cite{celebA}. Visual results corresponding to an upsampling factor of 4 is shown for FFHQ and Helen. For an upsampling factor of 8, we show the comparative results on CelebA and FFHQ. Methods used are Bicubic Interpolation, SRCN\cite{SRCNN}, EDSR\cite{EDSR}, FSRNet\cite{FSRNet}, DIC\cite{DIC} and SPARNet\cite{SPARNet}. Zoomed in images of left eye and mouth is shown to discern the quality of super-resolution. Further for differentiating the comparative perceptual quality of different methods, the image can be zoomed up to 10×.  }
\end{figure*}
\begin{figure*}
    \centering
    \includegraphics [scale=0.12] {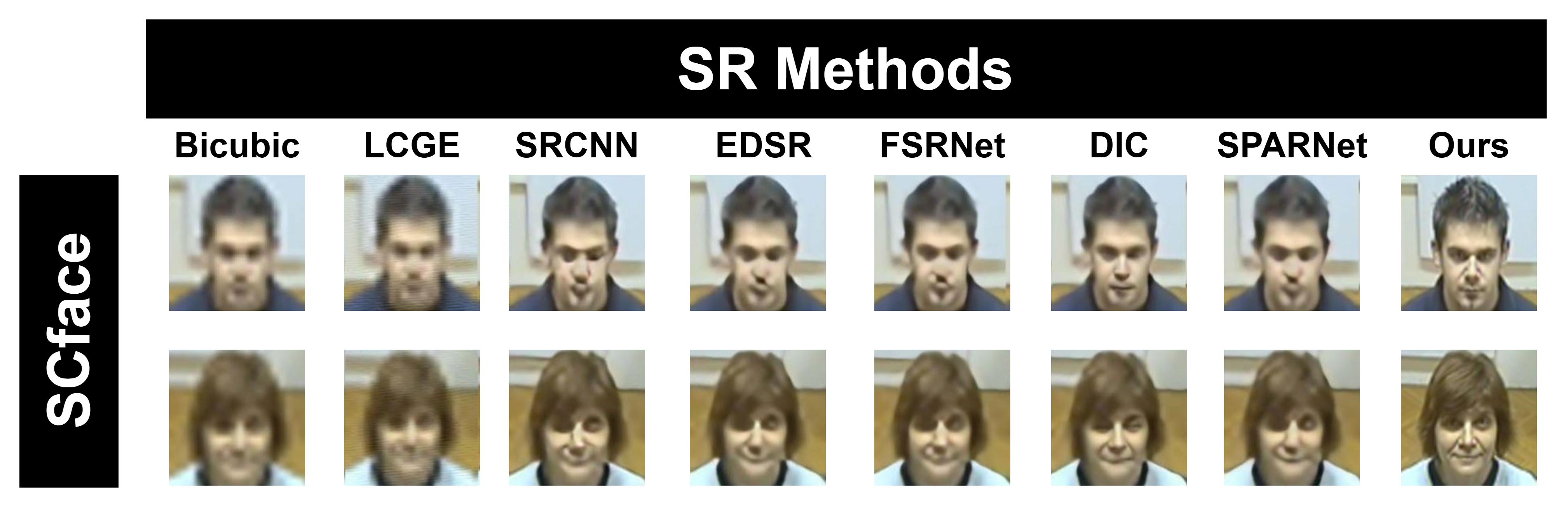}
    \caption{\label{vscface} Subjective visual performance on real-world surveillance scenarios for 8× SR, of SCface dataset. Visual comparisons are shown on two sample images from the dataset. }
   
\end{figure*}
\subsubsection{Qualitative Results}
Additionally, we provide visual comparisons of the output from different methods in Fig. \ref{visual}. While all approaches are able to reconstruct the basic facial structure, our method excels in restoring fine facial details, particularly key features like lips, teeth, and eyes. The preservation of the fine structure of face ensures robust retention of facial characteristics and expressions. This is achieved through the structural and spatial information embedded in the facial landmark features extracted via the GCN block.

The SRCNN method \cite{SRCNN} generally produces blurred results with distorted facial features, as seen in rows 6 and 8 of Fig.\ref{visual}. Both EDSR \cite{EDSR} and FSRNet \cite{FSRNet} tend to overly smooth the images, occasionally distorting certain facial regions, which is especially noticeable in rows 2, 6, and 8. Similarly, DIC \cite{DIC} results in a smoothed image with significant loss of facial texture, most evident in row 8. While SPARNet generally performs well in retaining overall image texture, it struggles significantly with certain facial features, such as the lips, teeth, dimples, and eyes. The lips show evident distortion in rows 1, 6, 7, and 9. The teeth and eyes are distorted in row 4. In rows 5 and 7, the dimples on the right cheek are completely lost. All these shortcomings are effectively addressed by our method.

We also present qualitative comparison results on two surveillance images from the SCface dataset \cite{SCFace}, clearly demonstrating that our method excels at preserving facial structures and maintaining accurate facial expressions in super-resolution tasks for low-resolution, real-world images.

In summary, both quantitative metrics and visual analysis confirm the superiority of our approach. 

\subsection{Ablation Study}
In this section, we further conduct experiments to verify the effectiveness of key components in AffectSRNet on ×8.
First, we try the network with only the super-resolution backbone with RRDB blocks (the network effectively becomes ESRGAN). Further InstanceNorm is added after every convolution layer in the RRDB blocks. Finally adding the GCN block and mergin its output in the super-resolution backbone gives our network. The metrics on each of the above networks are reported in Table \ref{ablnsr}. Introduction of InstanceNorm increases the metric marginally, this is in accordance to the fact that InstanceNorm performs well in image style transfer and related tasks\cite{InstanceNorm1, InstanceNorm2}. Finally, merging structural information from facial landmarks via GCN block, our method AffectSRNet achieves the best performance in terms of PSNR, SSIM and LPIPS and also preserves the facial expressions much better as reflected in the ECM values.

\section{Conclusion}
In conclusion, this research presents a novel approach to addressing the challenge of facial expression distortion during the upscaling of low-resolution images. By introducing AffectSRNET, an emotion-aware face super-resolution pipeline, we have demonstrated the feasibility of enhancing the resolution of facial images while preserving the integrity and intensity of facial expressions. This contribution has important implications for the practical application of facial expression recognition (FER) in real-world, low-resolution scenarios, such as video surveillance and human-computer interaction.

Moreover, by proposing a new metric to evaluate the accuracy of emotion-aware face super-resolution, this study offers a valuable tool for future research in this underexplored subdomain of facial super-resolution. The comprehensive evaluation of AffectSRNET against state-of-the-art FSR methods using established datasets, such as CelebA, FFHQ, and Helen, underscores the robustness of our approach and its potential for widespread adoption in FER systems.

This research lays the groundwork for further advancements in expression-aware FSR, and its findings can be extended to a variety of fields, including security, education, and entertainment. As deep learning techniques continue to evolve, the integration of emotion-aware solutions like AffectSRNET will play a pivotal role in enhancing the quality and effectiveness of FER applications, particularly in challenging low-resolution environments.

\section{Declarations}
\subsection*{Funding}
This research is supported by grants from Kwikpic AI Solutions.

\subsection*{Competing interests}
Financial interests: This research was supported by funding from Kwikpic AI Solutions. The authors declare no other competing financial or non-financial interests relevant to the content of this article.
\subsection*{Data and Code availability}
The datasets used for this study are publicly available and are cited in the references \cite{celebA, FFHQ, helen}. The authors would share the code upon a reasonable request.

\subsection*{Authors' contributions}
\begin{itemize}
    \item \textbf{Syed Sameen Ahmed Rizvi:} Conceptualization, Project Administration, Visualization, Writing - Original Draft, Writing - Review \& Editing.
    \item \textbf{Soham Kumar:} Investigation, Methodology, Investigation, Software, Visualization, Writing - Original Draft, Visualization.
    \item \textbf{Aryan Seth:} Conceptualization, Investigation, Methodology, Project Administration, Software, Validation, Writing - Original Draft, Writing - Review \& Editing.
    \item \textbf{Pratik Narang:}  Supervision, Writing - Review \& Editing and Resources.
\end{itemize}


{\small
\bibliographystyle{ieee}
\bibliography{egbib}
}

\end{document}